\documentclass[conference]{IEEEtran}
\IEEEoverridecommandlockouts

\usepackage{cite}
\usepackage{amsmath,amssymb,amsfonts}
\usepackage{algorithmic}
\usepackage{graphicx}
\usepackage{textcomp}
\usepackage{xcolor}
\usepackage{array}
\usepackage{multirow}
\usepackage{booktabs}
\usepackage{url}
\usepackage{hyperref}

\renewcommand{\arraystretch}{1.15}

\begin{document}

\title{Reducing the Complexity of Deep Learning Models for EEG Analysis on Wearable Devices%
\thanks{}%
}

  
\author{%
  \IEEEauthorblockN{%
    Farough Shayestehroodi\textsuperscript{1},
    Parham Zilouchian Moghaddam\textsuperscript{1},
    Mahdi Mohammadinasab\textsuperscript{1},\\
    Mehdi Modarressi\thanks{*Corresponding author, email:modarressi@ut.ac.ir}\textsuperscript{1*},
    Mostafa Ersali Salehi-nasab\textsuperscript{1},
    and Masoud Daneshtalab\textsuperscript{2,3}%
  }
  \IEEEauthorblockA{%
    \textsuperscript{1}School of Electrical and Computer Engineering, College of Engineering, University of Tehran, Tehran, Iran\\
    \textsuperscript{2}Intelligent Future Technologies (IFT), M\"alardalen University, V\"aster\aa{}s, Sweden\\
    \textsuperscript{3}Department of Computer Systems, Tallinn University of Technology (TalTech), Tallinn, Estonia\\
  }
}

\maketitle

\begin{abstract}
Wearable healthcare devices are the fastest-growing Internet of Things (IoT) sector. Many automated healthcare services rely on two crucial biological signals, namely ECG and EEG, which reflect the activity of the heart and brain, respectively. Although deep neural networks are considered the primary way to process and analyze these signals, the very tight energy and computational power constraints in wearable devices are far below the computational, energy, and memory bandwidth demands of DNN models, thereby impeding the deployment of deep learning in many practical wearable services. This paper investigates the feasibility of deploying state-of-the-art DNN models in resource-constrained wearable devices. Notably, we explore the trade-off between accuracy and computational complexity of DNNs when parameter quantization and electrode reduction methods are used. Our investigation centers on several state-of-the-art DNN models designed for EEG signal analysis, specifically for detecting epileptic seizures. Our findings demonstrate that, when applied judiciously, these techniques can significantly reduce the complexity of the DNNs under consideration with minimal adverse effects on accuracy. These results reveal the explicit trade-offs between accuracy and complexity reduction encountered when adapting DNN-based online EEG analysis for wearable devices.
\end{abstract}


\section{Introduction}

Wearable devices are intelligent electronic devices incorporated into clothing or worn on the body, as implants or accessories, to automatically monitor the user's vital signs and biological data. Perhaps automated healthcare is the primary application of wearables: the spread of wearable devices, in the form of the Internet of Medical Things (IoMT) \cite{alansari2018iot}, has made it possible to monitor health continuously, particularly for high-risk patients, changing episodic manual sampling approaches to continuous monitoring and automated intervention.

Analyzing some sensory data on wearable devices, such as heart rate or blood sugar level, requires quite simple calculations. Still, some vital and critical biological data, including the electroencephalogram (EEG) signal, are very complex to process. Today, Deep Learning (DL) is considered the premier approach for EEG analysis. DL is the most recent and promising trend in artificial intelligence, using Deep Neural Networks (DNNs) to learn to solve problems \cite{daneshtalab2020hardware}. There is a vast range of DNN models for EEG signal analysis that use different network architectures, such as convolutional neural networks (CNNs), Long Short-Term Memory (LSTM) recurrent networks, Transformer networks, or combinations of them, to classify EEG samples and detect patterns of interest.

An EEG signal is recorded by placing electrodes at specific locations on the scalp. It reflects the brain's electrical activities and is widely used to detect critical diseases and abnormalities of the nervous system. A healthy human EEG shows specific activity patterns, and any deviation from this normal pattern due to neurological diseases and disorders can be detected by EEG analysis \cite{huang2023epilepsynet}. Perhaps the most famous example is Epilepsy: it is one of the most common neurological disorders affecting more than 50 million people worldwide \cite{who2023epilepsy}. Epileptic seizures can be detected in advance by specific changes in the EEG signal through continuous monitoring of high-risk patients \cite{epilep}. Besides medical purposes, EEG analysis on wearables has diverse applications in other sectors, including EEG-based smart cars \cite{smartcar} and driver drowsiness detection, emotion recognition \cite{dep1}, and modern brain-computer interaction (BCI), to name a few \cite{bci1,bci2}.

Clinical EEG analysis is a piece of hospital equipment that has long been an essential component in brain disease diagnosis and assessment. Continuous real-time EEG monitoring relies on a lightweight, portable version of such equipment that integrates a sensor unit (a set of electrodes and amplifiers to capture signals) and a processing unit (to analyze the data or transmit it to a remote device) into a wearable device.

Deploying real-time EEG analysis on wearable devices, which are expected to achieve the same accuracy and performance as hospital equipment, is quite challenging. In particular, to enable 24/7 patient monitoring outside medical environments, wearable devices should use simple processing units with low battery consumption. In general, tight energy and resource constraints are common features of wearable devices. They are limited to using miniaturized processors and small coin-cell batteries to keep the system's battery lifetime, weight, and size reasonable, so as not to affect users' comfort. With these limitations, taking full advantage of complex DNN models on wearable devices is quite challenging: A moderately-sized DNN model for EEG processing requires several hundred thousand to millions of arithmetic operations to process a single time window of the EEG \cite{lee2022realtime}. While a clinical EEG monitoring device can deploy a powerful graphics processing unit (GPU) or a microprocessor (CPU) to run such algorithms in real time, this load is quite prohibitive for wearable devices. Many existing wearable devices operate under tight resource constraints by offloading compute-intensive tasks to a remote cloud computer and transmitting the collected data over a wireless connection for processing. However, due to some practical issues, including high power consumption of continuous wireless data transmission, loss of data privacy and security, loss of autonomy, and the need for an always-on wireless internet connection, there is always a strong demand for the so-called edge processing, in which EEG processing is done in (or close to) the wearable device to improve response speed, energy efficiency, and privacy. 
This paper studies the feasibility of deploying current state-of-the-art DNNs in resource-constrained systems. We focus on models that detect epileptic seizures by analyzing EEG signals. The paper's primary contribution is exploring the impact of low-bit-width quantization on these DNN models. Key findings include:

\textbf{Optimal Network Configuration:} We compare the accuracy of several state-of-the-art DNN models for epileptic seizure detection under wearable device conditions. Based on the results, a model that combines a CNN-based feature-extraction module and a back-end LSTM is the most effective at balancing accuracy and execution speed. We then continued the study with this selected model. We showed that this configuration can maintain output integrity when all feature maps and parameter data are reduced to 8-bit integer representations, with minimal error increase.

\textbf{Parameter binarization:} The paper explores the application of the state-of-the-art LQNet binarization method \cite{zhang2018lqnets}, which reduces the bit width of parameters and feature maps to 4 bits. The results show that the model retains functional efficacy even with this significant quantization.

\textbf{Impact of Electrode Reduction:} This paper also investigates how reducing the number of electrodes affects model performance. It successfully identifies the minimum set of electrodes required to maintain accuracy, providing valuable insights for optimizing EEG setups in wearable devices, where reducing the number of electrodes is critical for decreasing device complexity and increasing user comfort.

Note that this paper does not develop any new DNN model for EEG analysis; instead, it evaluates the performance of some existing methods when the proposed complexity reduction methods are applied.

The rest of the paper is organized as follows. Section~\ref{sec:background} reviews background material and introduces the considered DNN models for EEG analysis. Section~\ref{sec:quantization} presents the methods and results of model quantization, followed by electrode reduction in Section~\ref{sec:input_reduction}. Finally, Section~\ref{sec:conclusion} concludes the paper.

\section{Background and Related Work}\label{sec:background}

\subsection{DNNs for EEG}

\textbf{DNN for EEG analysis.} A survey on recent advances in DNN-based EEG analysis can be found in \cite{li2020deepeeg}. A recent work \cite{li2020deepeeg} evaluates some combinations of DNN models for EEG analysis. The results show that methods combining CNN (which captures the spatial characteristics of EEG) with RNN (which captures the temporal characteristics of EEG) achieve the highest accuracy. Transformer-based methods are the most robust. However, the primary goal of these methods is performance, and they do not consider the hardware limitations of wearable devices \cite{mirsalari2022factlstm}.

\textbf{DNNs in wearable devices.} Most existing edge-based wearable systems replace the more efficient DNN with the more straightforward (but less accurate) alternative methods for EEG \cite{oleary2020neuromorphic,wang2020closedloop,epfl2023eglass}. An example is the e-Glass project \cite{epfl2023eglass}, a wearable system that uses four EEG electrodes embedded in the patient's glasses to detect epileptic seizures. It uses a simple microcontroller to analyze data and, hence, does not employ a DNN; instead, it uses a more straightforward decision-tree method to detect abnormalities in EEG.

Whereas most DNN-based EEG processing methods target cloud-side systems, some recent papers have addressed the need for an edge-based DNN accelerator chip for biological signal analysis \cite{tsai2023scicnn}.

\subsection{Selected DNNs for EEG analysis}

This section introduces the most prominent networks considered while finding the base model. All models are taken from recent studies in \cite{lee2022arxivseizure} and \cite{roy2019chrononet}.

\begin{itemize}
  \item \textbf{CNN-LSTM:} uses a front-end ResNet network to extract features from the signal. ResNet is a compelling and accurate CNN model. The features are then fed to an LSTM network to generate meaningful information from the sequence of observed features.
  \item The \textbf{ChronoNet} network \cite{roy2019chrononet} uses several layers of filters on top of an RNN to detect particular patterns in the input signal.
  \item The \textbf{ResNet18} network is a ResNet-based model for processing signals as images.
  \item The \textbf{2D-CNN-LSTM} network differs from ResNet-LSTM's architecture by using 2D-CNN layers instead of ResNet. Its inclusion in the competition will help to determine the effects of using ResNet layers, which are deeper and have skip connections, on various aspects of the chosen model.
  \item In \textbf{ResNet-BiLSTM}, a bidirectional LSTM (called BiLSTM) replaces LSTM in the ResNet-LSTM structure.
  \item \textbf{ResNet-GRU} uses a front-end GRU instead of an LSTM.
\end{itemize}

\subsection{Comparison criteria}

We have compared our model with several other models using various criteria to demonstrate its capability and real-world performance. Below are the criteria we used for this purpose:

\begin{itemize}
  \item \textbf{AUROC:} The Area Under the Receiver Operating Characteristic (AUROC) is a metric to assess how well a binary classification model performs. It ranges from 0 to 1, with a higher value indicating a better ability to differentiate between two classes. The calculation involves plotting the True Positive Rate against the False Positive Rate at different thresholds and finding the area under the curve.
  \item \textbf{AUPRC:} The AUPRC is a metric that evaluates a binary classification model's performance, especially in imbalanced datasets. A higher value implies better performance. This metric measures the area beneath the curve that plots precision (positive predictive value) against recall (true positive rate) at different thresholds.
  \item \textbf{TPR:} The True Positive Rate (TPR), also referred to as sensitivity or recall, is a metric used in binary classification to determine the proportion of actual positive results that the model correctly identifies. It is calculated by dividing the number of true positives by the sum of true positives and false negatives.
  \item \textbf{TNR:} The True Negative Rate (TNR) is a metric used in binary classification to assess how well a model can identify actual negatives. It's also known as specificity. The TNR is calculated by dividing the number of true negatives by the sum of true negatives and false positives.
  \item \textbf{F1 Score:} The F1 score is a metric used in binary classification to evaluate the accuracy of a model. It combines both precision and recall into a single score. The score is calculated as the harmonic mean of precision (true positives divided by the sum of true and false positives) and recall (true positives divided by the sum of true positives and false negatives).
\end{itemize}
\subsection{Noise and artifacts}EEG signals are highly susceptible to various forms of degradation that complicate accurate data analysis. Broadly categorized, these distortions manifest as structural artifacts or stochastic noise. Physiological artifacts stem directly from the body activities, such as eye blinks and retinal dipoles (ocular), muscle contractions during swallowing or talking, and cardiovascular fluctuations (cardiac)  \cite{jiang2019removal}. On the other hand, extrinsic or environmental noise originates from external hardware and experimental setups, including power line interference (typically 50/60 Hz), high-frequency instrument noise, and electrode displacement transients \cite{noise2,jiang2019removal}. Because these overlapping contaminants can mimic or strongly mask subtle underlying neurological oscillations, a robust denoising process is essential to prevent misleading clinical interpretations or flawed downstream machine learning classifications \cite{kaya2019brief}.
Several different methods are applied to clean these signals. Common methods for removing artifacts include simple digital filters, regression methods, adaptive filters, Independent Component Analysis (ICA), and standard supervised deep learning models \cite{kaya2019brief}. Supervised deep learning models can learn to suppress such artifacts by training on noisy signals paired with their clean references. Reconstructing clean images from noisy inputs by training supervised generative deep learning models for noise removal is a well-established field of research \cite{vqe}. The seizure-detection models considered in this paper, however, are trained as classifiers directly on real, artifact-containing EEG rather than as denoisers; robustness to these artifacts is therefore acquired implicitly from the training data rather than through an explicit denoising stage. Previous work demonstrated that Neural Architecture Search (NAS) can enhance robustness against noisy data \cite{vahid}. For future work, we plan to apply a NAS procedure to design a DNN that optimizes not only task performance and execution time but also robustness.
\section{Model Quantization}\label{sec:quantization}

In this section, we compare some selected cutting-edge DNN models for EEG analysis to choose the model with the highest accuracy. Table~\ref{tab:performance} shows the accuracy results based on the evaluation criteria discussed earlier in Section~\ref{sec:background}.

Throughout the paper, we use the TUSZ dataset \cite{tuh2023tusz} to train and evaluate the networks, as it is the largest publicly available dataset with a wide variety of signals. The network's raw EEG signal input data contains 20 bipolar channels and is resampled at 200~Hz to lower the storage space requirement. The data is then divided into 4-second signal segments and used as the network input, as in previous studies. Therefore, one sample of the network input consists of 800 time steps and 20 bipolar channels (See Section~\ref{sec:input_reduction}).

The reported results are the average performance of the models across five different seeds. Using different seeds is a common method to reduce the impact of the initial starting point of the network weights on neural network performance. The dataset imbalance and the high cost of false-negative detection are the reasons for choosing AUROC and AUPRC as the primary evaluation criteria for selecting the best model. Throughout, unless otherwise noted, we report AUROC (\%) as the primary accuracy metric. According to Table~\ref{tab:performance}, it is evident that the ResNet-LSTM model has performed better than other models.

\begin{table*}[!t]
\caption{Performance Comparison of EEG-Analysis DNN Models on the TUSZ Dataset}
\label{tab:performance}
\centering
\renewcommand{\arraystretch}{1.25}
\setlength{\tabcolsep}{10pt}
\begin{tabular}{lccccc}
\toprule
\textbf{Model} & \textbf{F1} & \textbf{TNR} & \textbf{TPR} & \textbf{AUPRC} & \textbf{AUROC} \\
\midrule
ResNet18      & $69.65 \pm 0.76$ & $78.66 \pm 1.59$ & $76.02 \pm 2.19$ & $73.59 \pm 1.62$ & $84.91 \pm 0.78$ \\
ChronoNet     & $51.67 \pm 0.90$ & $65.09 \pm 6.63$ & $54.43 \pm 6.83$ & $45.53 \pm 2.79$ & $62.70 \pm 2.58$ \\
2D-CNN-LSTM   & $73.45 \pm 1.57$ & $82.07 \pm 1.33$ & $78.48 \pm 2.48$ & $80.54 \pm 2.37$ & $87.52 \pm 1.29$ \\
ResNet-GRU    & $76.39 \pm 1.45$ & $84.44 \pm 2.38$ & $\mathbf{80.60 \pm 2.38}$ & $83.00 \pm 1.66$ & $90.05 \pm 0.94$ \\
ResNet-BiLSTM & $74.38 \pm 1.66$ & $81.82 \pm 1.21$ & $80.31 \pm 2.30$ & $80.58 \pm 1.97$ & $88.60 \pm 1.44$ \\
\textbf{ResNet-LSTM} &
$\mathbf{76.93 \pm 1.35}$ & $\mathbf{85.30 \pm 1.17}$ & $80.55 \pm 3.01$ &
$\mathbf{83.93 \pm 1.85}$ & $\mathbf{90.40 \pm 1.01}$ \\
\bottomrule
\end{tabular}
\\[3pt]
{\footnotesize Values report mean $\pm$ standard deviation over five random seeds. Best result per column is in bold.\par}
\end{table*}

In Table~\ref{tab:exec_mem}, the parameters size and processing load are compared. The model parameters were measured using the \textit{torchinfo} tool. The number of arithmetic operations the model executes to process a single input (a sample window of the EEG signal) measures the processing load (multiply-and-accumulate operations). The results are the average execution load across 500 data samples (with a batch size of 1).

\begin{table}[!t]
\caption{Execution and Memory Profile Comparison}
\label{tab:exec_mem}
\centering
\renewcommand{\arraystretch}{1.25}
\setlength{\tabcolsep}{4pt}
\begin{tabular}{lcc}
\toprule
\textbf{Model} & \textbf{Parameters Size} & \textbf{Processing Load} \\
               & \textbf{(MB)}            & \textbf{(Giga Op.s)}    \\
\midrule
ResNet18              & 20.82          & 5.742          \\
ChronoNet             & \phantom{0}0.50  & 0.015          \\
2D-CNN-LSTM           & \phantom{0}6.16  & 0.285          \\
ResNet-GRU            & 14.70          & 0.983          \\
ResNet-BiLSTM         & 13.12          & 0.982          \\
\textbf{ResNet-LSTM}  & \textbf{15.75} & \textbf{1.153} \\
\bottomrule
\end{tabular}
\end{table}

\subsection{Model selection} 
\begin{figure*}[!t]
\centering
\includegraphics[width=\textwidth]{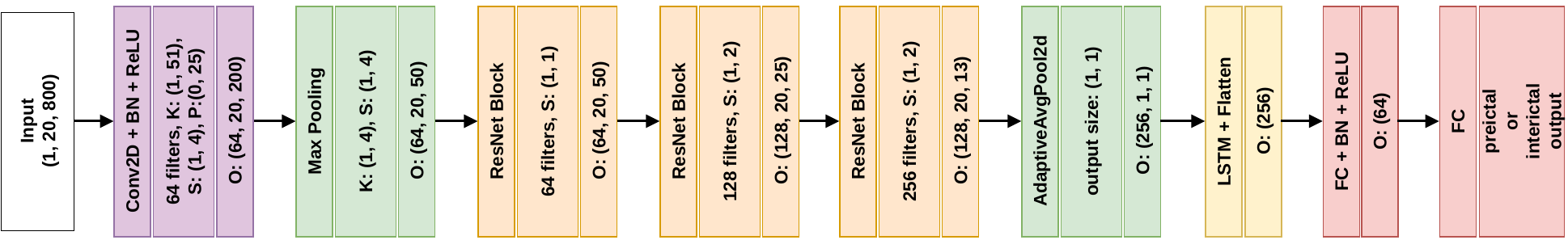}
\caption{The architecture of the ResNet-LSTM network.}
\label{fig:resnet_lstm}
\end{figure*}

From this point on, we will continue our experiments with the ResNet-LSTM model, which exhibits the best accuracy in EEG classification. Fig.~\ref{fig:resnet_lstm} shows the structure of this model.

This particular model typically consists of three main components: convolutional layers based on ResNet Blocks, an LSTM recurrent layer, and fully connected layers for binary classification. The architecture of this model is illustrated in Fig.~\ref{fig:resnet_lstm}, where K denotes the filter size, S indicates the stride, P represents padding, and O represents the output size of that layer.

The network starts with a 2-D CNN that extracts signal shape features along the time axis. This is followed by a MaxPooling layer, which helps reduce computation in subsequent layers while retaining important features. Then, three ResNet layers are used, which not only reduce the input size in the time dimension but also enable the model to learn more complex signal shapes and improve gradient flow. The convolution layers have one-dimensional kernels. When extracting information, it is important to consider the time axis and each channel separately. This approach ensures that any significant information recorded by one channel but not others is preserved. Additionally, it prevents information from different channels from interfering with one another.

\begin{figure}[!t]
\centering
\includegraphics[width=0.83\columnwidth]{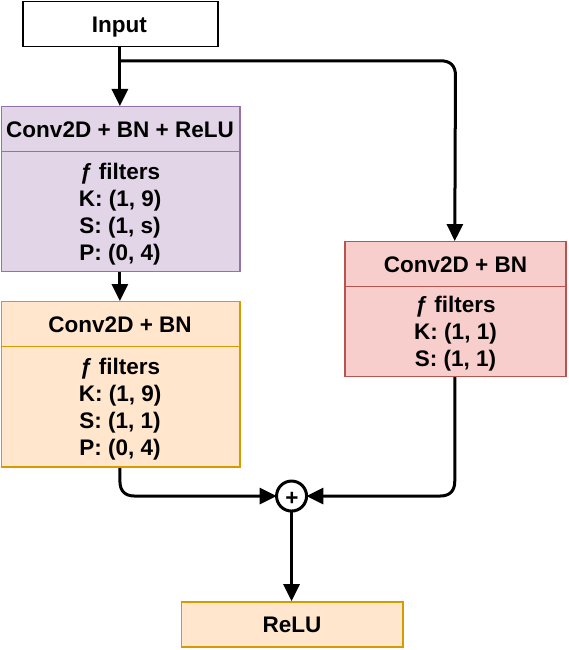}
\caption{The inner architecture of the ResNet Block.}
\label{fig:resnet_block}
\end{figure}

Fig.~\ref{fig:resnet_block} depicts the inner structure of ResNet Blocks. These blocks consist of three different 2D convolution layers. The structure contains two separate data flows that are eventually concatenated and fed into the ReLU activation function. Using one-dimensional kernels with two-dimensional convolution is computationally more efficient and faster than using multiple channel-wise CNN1Ds. It's worth noting that this particular model is quite flexible in the number of channels it can support. Essentially, altering the number of input data channels doesn't change the model's structure, so it can be used with a different number of electrodes or retrained as needed. This makes it an excellent choice for continuous learning applications. After the final ResNet-based layer, an AdaptiveAvgPooling layer is added. This layer calculates an average feature for each feature plane through two-dimensional averaging. This is done for several reasons: first, in the initial parts of the network, where the feature space is not yet rich and contains more noise or other insignificant or irrelevant information, it's common to use MaxPooling. Meanwhile, in the final parts, where the feature space contains essential information, AvgPooling is used. Secondly, the adaptability of the AdaptiveAvgPooling layer makes the model more robust to changes in the input data size along the time dimension. This means the network's input data can be used with different durations without altering the model's structure. Thirdly, it prepares the input for the next layer, LSTM. Finally, after applying AdaptiveAvgPooling and removing 1-dimensional dimensions, a 256-dimensional feature vector (the number of filters in the last ResNet layer) is obtained and fed to the LSTM layer to learn temporal features and dependencies. The output of the LSTM layer is then fed into the network's classification part, which consists of two fully connected layers.

\subsection{Parameter quantization} Quantization is the most effective mechanism for reducing the computation overhead \cite{daneshtalab2020hardware}. Neural network computations are typically performed using 32-bit floating-point numbers, which can slow the model and increase its size. To address this issue, model quantization (also known as model compression) is used. There are two types of quantization: Quantization Aware Training (QAT) and Post-Training Quantization (PTQ) \cite{pytorch2023quant}. QAT methods can be time-consuming, so we looked into PTQ methods available in PyTorch. However, these methods only support quantization from 32-bit floating-point (Float32) to 8-bit integer (Int8).

There are two types of post-training network quantization: static and dynamic. With static quantization (PTSQ), the network weights and activation functions are quantized, and the activation functions are combined with those of previous layers when possible. This method incurs no time overhead during model inference. With dynamic quantization (PTDQ), activation functions are quantized dynamically during inference, incurring some time overhead compared to PTSQ. It is important to note that PTSQ supports quantization of convolutional and fully connected layers, whereas PTDQ supports quantization of LSTM and fully connected layers. During our research, we examined the quantization of Resnet-LSTM network layers using both methods. However, it is essential to note that the results showed that PTSQ, despite good model compression, leads to a severe performance decline. On the other hand, PTDQ does not significantly degrade the model's performance, but only about 27\% of the network parameters are quantized with it.

We use PyTorch to reduce the bit width of the two parts of the system, i.e., the CNN and LSTM, from 32-bit floating-point to 8-bit integers. The results show that with 8-bit quantization of the input data and network parameters, accuracy drops significantly, falling below 48\%. To investigate the source of this inefficiency, we conducted two more experiments and applied the quantization to one part of the system at a time: first, the CNN's parameters are quantized to 8 bits, and LSTM keeps the original 32-bit parameters; then, LSTM is quantized, and the CNN is kept intact. The results show that the CNN is more sensitive to bit-width reduction: with an 8-bit LSTM and a 32-bit CNN, the model's accuracy is 90.38\%, approximately equal to the original. However, with a 32-bit LSTM and an 8-bit CNN, the accuracy drops to 60.8\%.

To explore the sensitivity of the CNN component to bit width, we evaluate the model's accuracy as the CNN bit width varies from 32 to 10 bits. In this experiment, the LSTM bit width is set to 8 bits.

\begin{figure}[!t]
\centering
\includegraphics[width=1.03\columnwidth]{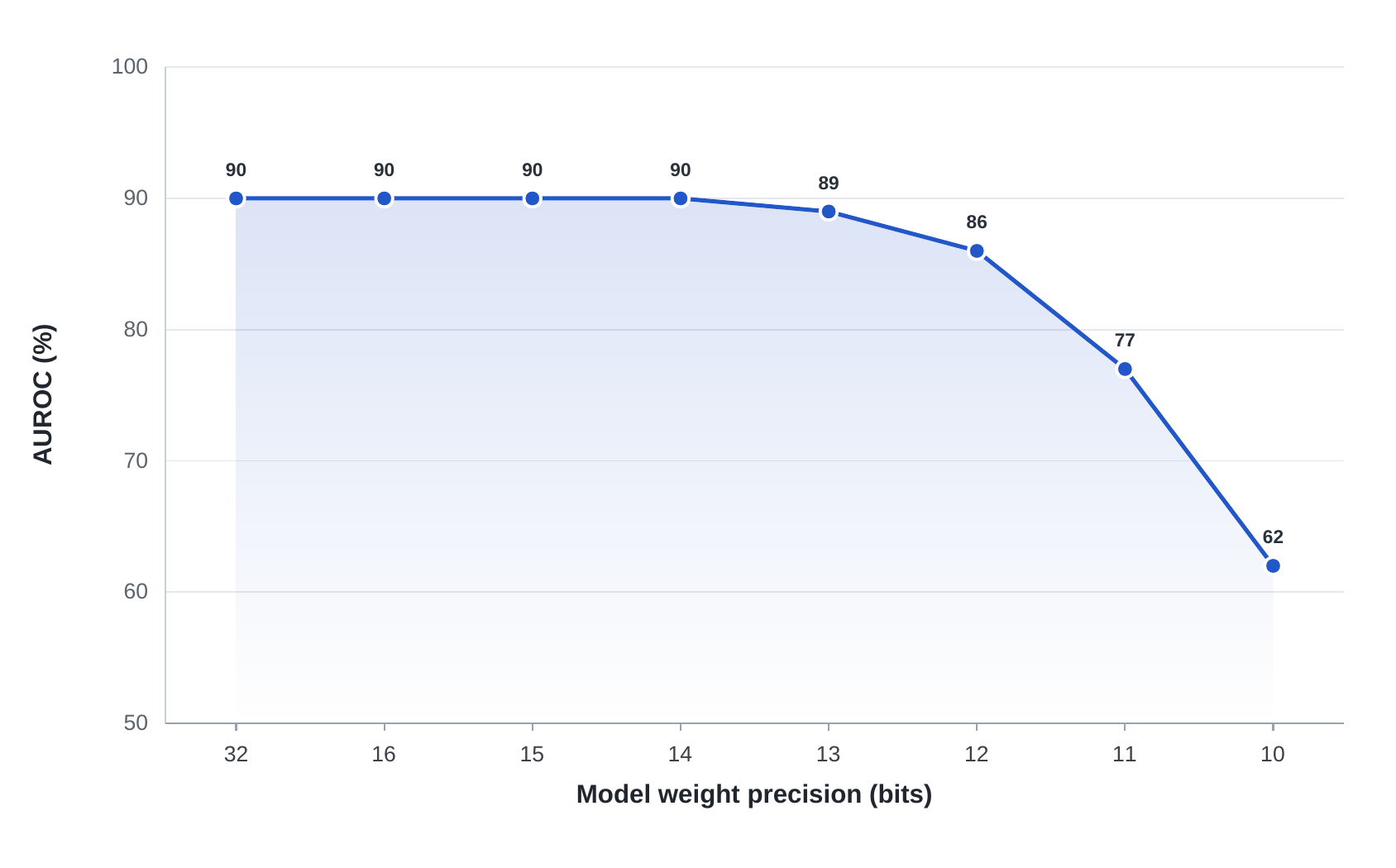}
\caption{The effect of CNN bit width on the model's accuracy.}
\label{fig:bitwidth}
\end{figure}

Fig.~\ref{fig:bitwidth} shows that, as we expected, the accuracy drops with bit width reduction. However, the accuracy loss is negligible up to 13 bits and then grows considerably. Thus, the 13-bit parameters are the most efficient option for the CNN part.

Note that the 13-bit parameter size applies only when specialized hardware is generated for the model; in this case, we can set the widths of registers, buses, and multipliers to 13 bits in a fine-grained manner to save space and bandwidth. However, when the model runs on a regular processor, which is very unlikely to support a 13-bit data width, we should use the nearest supported bit width, which is 16-bit for most processors.

Overall, with an 8-bit LSTM and a 16-bit CNN, the model's size is reduced from the original 15.75~MB (see Table~\ref{tab:exec_mem}) to 6.4~MB, a 59\% reduction.

Quantization does not change the number of arithmetic operations. Instead, it reduces the delay of each operation since we now perform simpler 8-bit and 16-bit integer operations rather than the baseline 32-bit floating-point ones. Several hardware-oriented methods leverage the baseline 32-bit multiplier to execute multiple lower-bitwidth operands in parallel (such as two 16-bit or four 8-bit operands). This approach effectively increases throughput when reduced-precision operations are required. \cite{negar}.  
When executing the original 32-bit model and the quantized model on the same processor, the results show that the quantized model's execution time is $3.4\times$ faster.

\subsection{Model binarization}

Binarization is an extreme case of quantization that constrains each weight to one of two values, typically $-1$ and $+1$. Although it has been shown to strike a good trade-off between accuracy and computational cost/model size in some networks, our study shows that it cannot be applied to our model, as it reduces accuracy below an acceptable level. Multi-level Binarization (MLB) in neural networks extends traditional binary networks by adding multiple binary layers with varying significance \cite{mbert,lin2017towards}. This enhances network capacity while retaining the simplicity of binary operations, thereby improving accuracy without significantly increasing complexity. LQ-Net \cite{zhang2018lqnets}, the most state-of-the-art MLB, further refines MLB by adjusting weights and activations, recognizing that a uniform quantizer may not yield the optimal solution. It advocates non-uniform binarization intervals, assigning varying levels of precision to network parameters.

By adjusting the binary level during training, LQ-Net can achieve higher accuracy than the traditional uniform post-training quantization method we have used to date in this paper, thereby enabling greater parameter bit-width reduction. LQ-Nets have demonstrated superior accuracy compared with previous methods, particularly on datasets such as ImageNet. In this paper, we implemented our networks on top of the LQ-Net quantizer. The results of implementing the method with a 4-level LQ-Net, compared to the network with the best conventional quantization (8-bit LSTM, 13-bit CNN), show that:

\begin{itemize}
  \item The model size is reduced from 6.4~MB to 2.3~MB.
  \item The execution speed is reduced by $4.1\times$.
  \item The accuracy (AUROC) is reduced from 90.33 to 86.12.
\end{itemize}

The results show that binarization significantly reduces execution time and memory footprint, but at the cost of a 5\% reduction in accuracy. If a device manufacturer decides this accuracy loss is still acceptable, it can benefit from the superior execution profile of the binarized version of our model.
\subsection{Further acceleration opportunities}
In this paper, we explore the trade-off between model complexity reduction, accuracy, and execution speed. This research will be extended in future work to identify the optimal model architecture and subsequently design specialized hardware, such as a custom chip for wearable devices. The inherent resilience of these models to quantization paves the way for computation reuse methods, which can considerably reduce both computational overhead and power consumption \cite{hoda,ali,ucnn}. Specifically, when low-bit quantization is applied, the vast set of model parameters is mapped to a highly restricted set of discrete values. This drastically increases the number of weight repetitions, thereby unlocking significant opportunities to bypass repetitive computations. Furthermore, the repetitive nature of the input data provides an additional avenue to eliminate redundant computations associated with repeated input signals \cite{mary,gonzalez,sara}. As demonstrated in our previous work, this dual approach has a significant positive impact on the energy efficiency of ECG processing \cite{mary}.
\section{Input Size Reduction}\label{sec:input_reduction}

\begin{figure}[!t]
\centering
\includegraphics[width=\columnwidth]{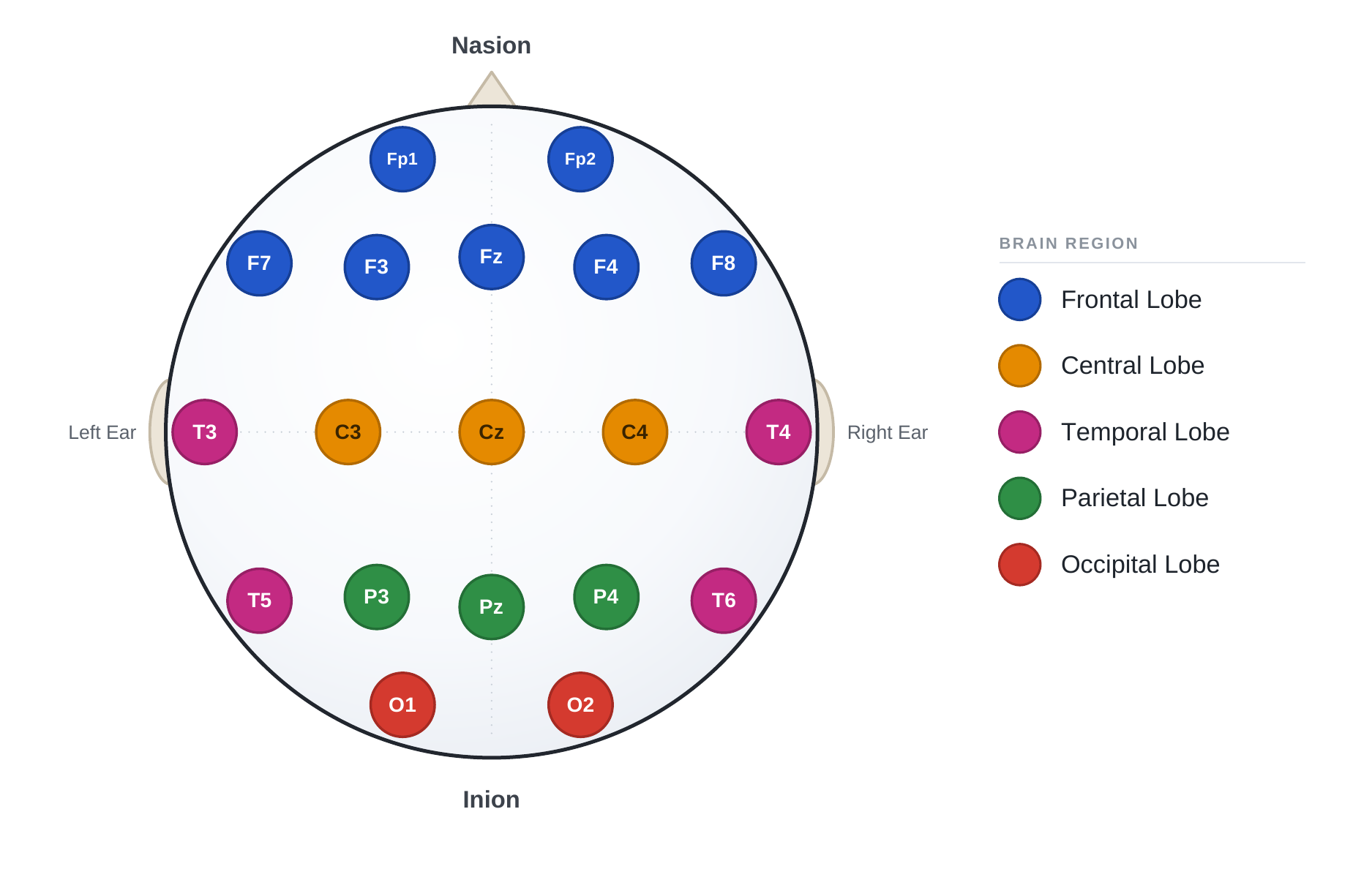}
\caption{EEG montage map in 10-20 standard \cite{liang2020scalpeeg}.}
\label{fig:montage}
\end{figure}

EEG signals, which record the brain's electrical activity, are typically collected using several electrodes placed on the skull. These electrodes, positioned at different locations on the skull, gather signals from various brain regions. The arrangement of these electrodes is referred to as a montage. In the case of an epileptic seizure, since different regions of the brain can be the origin of seizure attacks, the montage used must provide adequate coverage of these various brain areas. This is typically done using the so-called 10-20 standard \cite{liang2020scalpeeg}. The system is named 10-20 because it uses a combination of 10\% and 20\% rules to position the electrodes, ensuring even coverage and consistent relative positioning between electrodes. In this system, electrodes are named with letters and numbers. The letters refer to the brain area under the electrode (`F' for the frontal lobe, `C' for the central lobe, `T' for the temporal lobe, `P' for the parietal lobe, and `O' for the occipital lobe). Odd numbers indicate electrodes on the left hemisphere, even numbers for the right, and `z' (zero) for midline electrodes. Clinical settings for epileptic seizure detection usually comprise 21 electrodes, and Fig.~\ref{fig:montage} illustrates this system. In research studies, especially those requiring precise localization of brain activity, systems with 128, 256, or more electrodes may be used.

On the other hand, portable devices used for long-term monitoring outside clinical settings employ fewer electrodes for practicality and patient comfort. In summary, the number of electrodes in an EEG setup can range from 20 in standard clinical settings to over 200 in specialized research contexts. The choice depends on the required resolution and specific clinical or research needs.

Epileptic seizure detection uses a bipolar montage, in which the data from two electrodes are treated as a single channel, and the voltage difference between them is measured and fed into the DNN for classification. Clinical setups use 20 electrodes for epileptic seizures. Still, when a deep learning method is used, fewer electrodes may capture seizure patterns due to redundancy among electrodes. To study the feasibility of reducing the electrodes, we select different combinations of electrodes/channels. All combinations should include the electrodes most sensitive to the epileptic signals \cite{liang2020scalpeeg}. Table~\ref{tab:electrodes} shows the results. In this table, the network is trained on data from all electrodes, but during inference, only data from the selected electrodes are fed into the network. Table~\ref{tab:electrodes} demonstrates that two arrangements in rows 7 and 12 (identified by bold font in Table~\ref{tab:electrodes}) have shown acceptable performance. Both arrangements have eight channels recorded by ten electrodes. This reduction from 20 to 8 channels indicates that, despite having only 40\% of the input data compared to the original setting, the model performs well and demonstrates excellent resilience to data loss. Reduced data volume lowers storage requirements and increases processing speed, since the weights and associated computations are no longer needed. Moreover, reducing the number of electrodes simplifies wearable device design, making them more comfortable for patients.

\begin{table}[!t]
\caption{Reduced Electrode Setups and Their Accuracy}
\label{tab:electrodes}
\centering
\footnotesize
\renewcommand{\arraystretch}{1.15}
\setlength{\tabcolsep}{4pt}
\begin{tabular}{@{}>{\raggedright\arraybackslash}p{1.45cm}
                  >{\raggedright\arraybackslash}p{4.0cm}
                  >{\centering\arraybackslash}p{1.80cm}@{}}
\toprule
\textbf{No.\ of channels (electrodes)} &
\textbf{Selected channels / electrodes} &
\textbf{AUROC} \\
\midrule
20 channels (19 electrodes) &
Fp1-F7, Fp2-F8, F7-T3, F8-T4, T3-T5, T4-T6, T5-O1, T6-O2, T3-C3, C4-T4, C3-Cz, Cz-C4, Fp1-F3, Fp2-F4, F3-C3, F4-C4, C3-P3, C4-P4, P3-O1, P4-O2 &
$90.39 \pm 1.01$ \\
\midrule
2 channels (4 electrodes) & Fp2-F8, T4-T6                 & $74.65 \pm 1.38$ \\
2 channels (4 electrodes) & F7-T3, F8-T4                  & $77.35 \pm 0.93$ \\
\midrule
4 channels (5 electrodes) & T3-T5, T4-T6, T5-O1, T6-O2    & $84.93 \pm 0.55$ \\
4 channels (5 electrodes) & T3-C3, C4-T4, C3-Cz, Cz-C4    & $85.28 \pm 0.77$ \\
\midrule
8 channels (10 electrodes) &
Fp1-F7, Fp2-F8, F7-T3, F8-T4, T3-T5, T4-T6, T5-O1, T6-O2 &
$86.25 \pm 1.05$ \\
\textbf{8 channels (10 electrodes)} &
\textbf{T3-T5, T4-T6, T5-O1, T6-O2, T3-C3, C4-T4, C3-Cz, Cz-C4} &
$\boldsymbol{89.23 \pm 0.72}$ \\
8 channels (9 electrodes) &
T3-C3, C4-T4, C3-Cz, Cz-C4, Fp1-F3, Fp2-F4, F3-C3, F4-C4 &
$85.95 \pm 0.86$ \\
8 channels (9 electrodes) &
Fp1-F7, Fp2-F8, F7-T3, F8-T4, T3-C3, C4-T4, C3-Cz, Cz-C4 &
$85.80 \pm 0.82$ \\
8 channels (12 electrodes) &
Fp1-F7, Fp2-F8, F7-T3, F8-T4, T3-T5, T4-T6, F3-C3, F4-C4 &
$83.58 \pm 0.97$ \\
8 channels (12 electrodes) &
Fp1-F7, Fp2-F8, T3-C3, C4-T4, F3-C3, F4-C4, C3-P3, C4-P4 &
$85.55 \pm 0.91$ \\
\textbf{8 channels (10 electrodes)} &
\textbf{T3-T5, T4-T6, T5-O1, T6-O2, T3-C3, C4-T4, P3-O1, P4-O2} &
$\boldsymbol{89.68 \pm 0.71}$ \\
\midrule
4 channels (unipolar) & F7, F8, T3, T4                 & $81.16 \pm 1.11$ \\
6 channels (unipolar) & Fp1, Fp2, F7, F8, T3, T4       & $80.99 \pm 1.26$ \\
6 channels (unipolar) & F7, F8, T3, T4, T5, T6         & $82.15 \pm 1.19$ \\
8 channels (unipolar) & Fp1, Fp2, F7, F8, T3, T4, T5, T6 & $82.07 \pm 1.24$ \\
\bottomrule
\end{tabular}
\\[3pt]
{\footnotesize The two best 8-channel/10-electrode configurations are shown in bold.\par}
\end{table}

\subsection{Execution profile with quantization and electrode reduction}

\begin{table*}[!htbp]
\caption{Execution and Memory Profile Comparison with Electrode Reduction and Quantization}
\label{tab:final_exec}
\centering
\renewcommand{\arraystretch}{1.25}
\setlength{\tabcolsep}{6pt}
\begin{tabular}{@{}>{\raggedright\arraybackslash}p{4.2cm}
                  >{\raggedright\arraybackslash}p{4.2cm}
                  ccccc@{}}
\toprule
\multicolumn{2}{c}{\textbf{Configuration}} &
\multicolumn{5}{c}{\textbf{Results}} \\
\cmidrule(lr){1-2}\cmidrule(lr){3-7}
\textbf{Bit width} & \textbf{Input} &
\textbf{AUROC} & \textbf{F1} & \textbf{FPR} &
\textbf{Model Size (MB)} & \textbf{Exec.\ time} \\
\midrule
Floating point 32-bit &
20 channels, 32-bit floating-point data &
90.39 & 76.9 & 14.70 & 15.75 & 1.00 \\
CNN: 13-bit integer, LSTM: 8-bit integer &
8 channels, 16-bit integer data &
88.60 & 73.8 & 19.20 & \phantom{0}5.73 & 0.22 \\
\bottomrule
\end{tabular}
\\[3pt]
{\footnotesize ``Exec.\ time'' is normalized to the 32-bit floating-point baseline.\par}
\end{table*}

Finally, we apply all the simplification methods to evaluate the model's accuracy and execution profile. The final model uses eight channels and ten electrodes (See Table~\ref{tab:electrodes}), 16-bit input data and intermediate feature maps, 13-bit CNN weights, and 8-bit LSTM and FC weights. Table~\ref{tab:final_exec} highlights the accuracy, model size, and normalized execution speed. The table shows:

\begin{itemize}
  \item A 61\% reduction in network size, which is the result of custom quantization for each network layer and removing the weights of the first layer related to the removed input channels.
  \item An 80\% reduction in the network input size, of which 60\% is due to the decrease in the number of channels and 20\% is due to the reduction in bit width.
  \item The combined improvements of 1 and 2 have resulted in at least a 70\% acceleration in network interpretation time.
  \item A decrease of less than 1\% in the AUROC metric and a 2.1\% decrease in the FPR metric.
\end{itemize}

\section{Conclusion}\label{sec:conclusion}

In conclusion, this paper presents a comprehensive study on deploying state-of-the-art Deep Neural Networks (DNNs) in resource-constrained wearable devices for epileptic seizure detection through EEG signal analysis. Key contributions include identifying an optimal network configuration that combines CNN and LSTM for balanced accuracy and speed, effectively implementing low-bit-width quantization with minimal error increase, and exploring the impact of electrode reduction on model performance. The methods reduce the model size by $3\times$ and execution time by $5\times$ with less than 1\% accuracy reduction. This research provides significant insights into optimizing DNN models for EEG analysis in wearables, balancing computational demands with the constraints of portable devices. These findings are crucial for advancing automated healthcare, particularly in continuous, real-time monitoring of neurological conditions, enhancing the efficiency and practicality of wearable technology in medical applications.



\end{document}